\pdfoutput=1
\documentclass[11pt]{article}

\usepackage[final]{acl}

\usepackage{times}
\usepackage{latexsym}

\usepackage[T1]{fontenc}
\usepackage[utf8]{inputenc}

\usepackage{microtype}

\usepackage{inconsolata}

\usepackage{graphicx}

\usepackage{todonotes}

\title{We Need to Measure Data Diversity in NLP --- Better and Broader}

\author{Dong Nguyen \\
  Utrecht University \\
  The Netherlands \\
  \texttt{d.p.nguyen@uu.nl} \\\And
  Esther Ploeger \\
  Aalborg University \\
  Denmark \\
  \texttt{espl@cs.aau.dk} \\}

\begin{document}
\maketitle
\begin{abstract}
Although diversity in NLP datasets has received growing attention, the question of how to \textit{measure} it remains largely underexplored. This opinion paper examines the conceptual and methodological challenges of measuring data diversity and argues that interdisciplinary perspectives are essential for developing more fine-grained and valid measures.
\end{abstract}

\section{Introduction}

 Language models seem to exhibit remarkable natural language understanding and generation capabilities. Yet numerous studies have pointed towards serious flaws:  they encode societal biases~\citep{gallegos-etal-2024-bias,hofmann2024ai} and are sensitive to spurious correlations \citep{10.1145/3596490}.
Many of these weaknesses have been traced back to the training data \citep{feng-etal-2023-pretraining,pnas_embers}. 
Moreover, as the size of these training datasets continues to increase, improvements in task performance show diminishing returns \citep{tirumala-etal-2023}.
Data size is often prioritized at the expense of data quality \citep{PAULLADA2021100336}.
However, attention for the quality and composition of NLP datasets is growing \citep{albalak2024a,elazar2024whats,van-noord-etal-2025-quality,yu-llms-2024}.

One  aspect is the diversity of datasets, i.e., \textbf{data diversity}, 
which plays a key
role in learning and generalization in both machine and human learning \citep{raviv2022variability}. 
Smaller, more diverse datasets have enabled competitive model performance despite their size \citep{zhou2023lima,schiller-etal-2024-diversity}.
Data diversity is also important for the fairness of models \citep{santy-etal-2023-nlpositionality}; for example, models trained on primarily standard language perform worse on dialectal data \citep{blodgett-etal-2018-twitter}.
For evaluation, limited diversity in test data can be a cause for poor performance estimates \citep{freitag-etal-2020-bleu}
and hinders the ability to distinguish between different systems \citep{sugawara-etal-2022-makes}.

Unfortunately, the concept of ``data diversity'' is challenging to define and measure, and researchers
are rarely explicit about what they mean with data diversity \citep{icml2024_position_paper}. 
But without clear definitions and measurements, it is difficult to make progress.
Previous papers that advocate for measuring data diversity have left important aspects unaddressed. \citet{mitchell2023measuringdata} emphasize the importance of measuring data characteristics, though diversity in NLP data was not their main focus. \citet{icml2024_position_paper}  highlight the need to measure data diversity; however, their discussion mostly focuses on considerations during dataset construction, rather than diversity measurement itself. \citet{shaib2025standardizingmeasurementtextdiversity} call for standardizing text diversity measurements; however, their work is mostly empirical and focuses on  lexical diversity.
\citet{esteve2025surveydiversityquantificationnatural} recently surveyed approaches to measure diversity in NLP; 
however, they focused less on data diversity, and 
they did not focus on
the broader challenges that we outline.
Further, the interdisciplinary focus of these papers is limited. 

Yet, data diversity measurement has long been of central importance in a wide range of fields,
including language learning and assessment \citep{scott2013}, corpus linguistics \citep{kilgarriff,Egbert_Biber_Gray_2022},
 sociology \citep{doi:10.1146/annurev-soc-030420-015435}  and  ecology \citep{simpson1949measurement}.
Across these and other fields, scholars have also grappled with how to define and quantify diversity. Our position is that an interdisciplinary approach is essential for improving data diversity measurement in NLP. 

\paragraph*{Contribution} In this opinion paper, we provide a critical perspective on the fundamental challenge of \textbf{measuring dataset diversity in NLP}.
We focus on language data and 
associated information (e.g., about the authors or annotators). We operationally define data diversity as \textit{the variability of the data across specific
dimensions (e.g., semantic, syntactic, lexical, sociolinguistic, genre, language, etc.)}.
Cross-lingual diversity has received increased attention in NLP \cite{joshi-etal-2020-state,ploeger-etal-2024-typological}, leading to standardized language diversity metrics \citep{ponti-etal-2020-xcopa,samardzic-etal-2024-measure}.
However, since within-language diversity remains underexplored, we focus on measuring diversity in monolingual settings. Still, our recommendations are not specific to any language. 
We first motivate \textit{why} data diversity should be measured  (\S\ref{sec:why}).
Then, we structure our discussion around three core challenges: diversity dimensions (\S\ref{sec:dimensions}), their measurement (\S\ref{sec:stat_properties}), and evaluation of measurement (\S\ref{sec:evaluation}).
Throughout our paper we highlight connections with other fields. Rather than offering definitive answers, we hope to spark discussion and to encourage more rigorous and interdisciplinary approaches to measuring data diversity.
 We conclude with concrete recommendations for NLP research that uses or develops diversity measures (\S\ref{sec:conclusion}).

\paragraph*{Related concepts}
 Related to data diversity is a dataset's \textit{representativeness}, which is widely discussed in corpus linguistics but still lacks a consensus definition \citep{Egbert_Biber_Gray_2022}. In other fields, like statistics and fairness in machine learning, the term has also been used in many ways \citep{Chasalow_facct}. One common view in corpus linguistics is that a representative corpus is a miniature of  the target population's language; some other views align more with how we define data diversity \citep{Egbert_Biber_Gray_2022,Stefanowitsch2020}. 

Another related concept is data \textit{bias}, which refers to systematic distortions in the data \citep{Olteanu2019}. It often refers to societal bias, e.g., when the data reflects historical prejudices, or statistical bias, e.g., due to issues like sampling biases and measurement errors \citep{annurev-statistics-042720-125902}. 
Although data bias and data diversity are related, they are not the same. For example, reducing statistical bias can increase diversity (e.g., by including more social groups), and some efforts to increase data diversity (e.g., oversampling minority groups) could increase statistical bias.

\section{Why Measure Data Diversity?}
\label{sec:why}

Although many papers make qualitative judgements about the diversity of datasets (e.g., \citealt{zhou2023lima,rottger-etal-2022-data}), the exact nature and extent of the differences in diversity can remain unclear. We take the view that data diversity should be \textbf{measured} --- that is, quantitatively characterized --- along different dimensions \citep{lai-etal-2020-diversity,mitchell2023measuringdata,icml2024_position_paper}. Measuring data diversity is relevant for both creating and reusing existing datasets, whether they are task-specific or more general-purpose.

\paragraph*{Measuring properties of datasets is essential for advancing NLP as a \textit{science}}
In \citeyear{kilgarriff}, \citeauthor{kilgarriff} wrote: ``\textit{In any science, one expects to find a useful account of how its central constructs are taxonomised and measured, and how the
subspecies compare. But to date, corpus linguistics has measured corpora
only in the most rudimentary ways, ways which provide no leverage on the
different kinds of corpora there are.}''  (p. 97).  This critique is also relevant to NLP: Quantitative analysis of dataset-level properties has been underexplored, with exceptions of recent work like \citet{lohr-hahn-2023-dopa} and \citet{ramponi-etal-2024-variationist}. 
Among these properties, diversity has been highlighted in the literature \citep{lai-etal-2020-diversity,mitchell2023measuringdata}.

\paragraph*{Measuring data diversity has the potential to improve every stage of the NLP pipeline}

First, it would support more deliberate and informed data collection and curation \citep{rogers-2021-changing}.
Data collection, especially when requiring human annotation, can be resource-intensive. Recent work \citep{quattoni-carreras-2021-minimizing-annotation, su2022selective,shi-etal-2021-diversity,alcoforado-etal-2024-random} shows that human annotation can be employed more effectively when selecting diverse datapoints.
Furthermore, quantitative metrics on dataset diversity 
could improve the documentation of new and existing datasets \citep{10.1145/3594737,gebru2021datasheetsdatasets} and contribute to dataset analysis tools \citep{lohr-hahn-2023-dopa,ramponi-etal-2024-variationist}.
This could support early intervention, during data collection, instead of post hoc. For example,
researchers could proactively collect more data from certain language varieties to improve the fairness of models.
Second, leveraging data diversity measurements for training and prompting language models  has led to better performing models, exhibiting 
improved robustness \citep{bukharin-etal-2024-data} and generalization  \citep{levy-etal-2023-diverse}.
Lastly, when it comes to evaluation, insufficiently substantiated diversity claims have caused disagreement on the generalizability of results.
For example, an evaluation dataset that covers ``a diverse range of text genres'' \citep{wang-etal-2018-glue} may, by other standards, deemed ``hardly diverse'' \citep{raji-etal-2021}.
Systematic data diversity measurement could help researchers to assess and interpret the generalizability of evaluation results.

\paragraph*{Measuring data diversity can clarify its impact and help explain and predict model behavior}
Current knowledge of data diversity in NLP is fragmented and coarse-grained, partly because it is often not measured or measured inconsistently across studies.
Rigorous data diversity measurement could reveal which dimensions of data diversity most strongly impact model behavior (e.g., task performance, generation quality, robustness, fairness), and how its effects interact with dataset size and other properties.
For example, \citet{guo-etal-2024-curious} study the effect of models being continuously trained on generated text
through the lens of data diversity. 
Besides explaining model behavior, diversity measurements could also serve as predictive signals.
For example, they  could help predict a model's task performance \citep{xia-etal-2020-predicting}, enabling researchers to better select datasets that align with their use cases, or anticipate fairness issues.

\section{Challenges in Measuring Data Diversity}
This section reflects on three key challenges when measuring data diversity:
(i) What diversity dimensions should we consider? (\S\ref{sec:dimensions});
(ii) How should we measure them?  (\S\ref{sec:stat_properties}); and
(iii) How should we assess the quality of the measures? (\S\ref{sec:evaluation}).

\subsection{Dimensions of Data Diversity}
\label{sec:dimensions}

NLP studies have considered a range of diversity dimensions. The most common are lexical diversity \citep{gandhi-etal-2024-better,guo-etal-2024-curious,lohr-hahn-2023-dopa,shaib2025standardizingmeasurementtextdiversity} and semantic diversity \citep{lai-etal-2020-diversity,li-etal-2024-scalingfilter,yu-etal-2022-data}. 
Studies have also considered various other broad dimensions, including syntactic \citep{guo-etal-2024-curious},  topical \citep{schiller-etal-2024-diversity}, and genre diversity \citep{liu-zeldes-2023-cant}, as well as more task-specific dimensions, including question types \citep{kim-etal-2023-aint,ghazaryan-etal-2025-syndarin}, key points in essays \citep{padmakumar2024does} and summarization styles \citep{grusky-etal-2018-newsroom}.

\paragraph*{A broader view: Consider multiple diversity dimensions}
The diversity of NLP datasets can be conceptualized along many dimensions. 
However, most studies focus on only one or at most two, with a few exceptions (e.g., \citealt{guo-etal-2024-curious}). 
This narrow focus risks  limiting our understanding of what constitutes a truly diverse dataset. Considering multiple dimensions simultaneously can also reveal how they correlate.  For instance, semantic diversity and lexical diversity may correlate considerably,  as expressing a broader range of meanings typically requires a more varied vocabulary. Some dimensions may also be better viewed as multidimensional. 
For example, using an umbrella term like ``instruction diversity'' while only measuring semantic diversity of instructions \citep{bukharin-etal-2024-data} risks that the results are interpreted more broadly than what the metric actually captures.

We cannot --- and need not --- measure all possible dimensions; likewise, datasets cannot (and need not) be highly diverse along all of them. 
Instead, we should selectively focus on the dimensions most relevant to the task at hand.
With this broader, yet targeted approach to data diversity measurement, we can clarify which dimensions matter for which tasks, and make more informed decisions about dataset construction.

\paragraph*{A broader view: Social dimensions of diversity}
Although NLP has increasingly considered various dimensions of data diversity, there remains a need to pay more attention to the social dimensions of diversity---especially those related to the identities of people who produce or annotate data. These aspects are central to the fairness of NLP systems, yet they get repeatedly ignored \citep{santy-etal-2023-nlpositionality}.

First, research should consider the sociolinguistic dimensions of diversity. Data collection and curation practices tend to favor \textit{standard varieties} of  languages  \citep{blodgett-etal-2020-language,gururangan-etal-2022-whose,nguyen2025}. 
Language, however, provides a multitude of ways (e.g., words, grammatical constructions) to express similar ideas, and this variation is shaped by social identity and context \citep{meyerhoff2018introducing}. 
NLP models primarily trained on standard language varieties often have lower performance on texts written in other language varieties or produced by speakers from underrepresented  groups \citep{faisal-etal-2024-dialectbench}.
Yet measuring diversity along dimensions such as stylistic \citep{wegmann-etal-2022-author} or dialectal \citep{grieve2025sociolinguistic} variation remains underexplored.

Second, research should also consider the background of a dataset's authors or annotators. Who produced or labeled the dataset will substantially influence what is included, excluded, or emphasized, with implications for fairness and bias in downstream models.
For instance, only modeling the majority vote of annotators can exclude minority perspectives, on both highly subjective tasks like hate speech detection and  seemingly more objective tasks like NLI \citep{sap-etal-2022-annotators,Cabitza_Campagner_Basile_2023,santy-etal-2023-nlpositionality}. Annotator diversity is also better seen as multidimensional; besides social categories like race or age, researchers have highlighted the importance of lived experience for annotator diversity \citep{10.1145/3544548.3580645}.

\paragraph*{A broader view: Consider the situational factors of texts}
NLP could draw from corpus linguistics by measuring the distribution of situational factors \citep{register_variation}, such as the mode of communication (spoken or written, synchronous or asynchronous). More broadly, diversity could be measured over registers, e.g., fiction, news articles, and academic writing \citep{annurev_register}. However, in NLP datasets, such information is often not available, and would need to be additionally annotated or inferred \citep{egbert_core}.

\subsection{How Should We Measure Diversity?}
\label{sec:stat_properties}
A single diversity dimension can often be measured in many ways. Consider semantic diversity, which is typically measured by first mapping text to embeddings \citep{lai-etal-2020-diversity,li-etal-2024-scalingfilter,yu-etal-2022-data}, followed by computations, like  the average pairwise similarity between sentences  or the entropy over clusters
\citep{guo-etal-2024-curious,han-etal-2022-measuring}. How the measure is formulated can influence both the interpretation of results and how the measure can be used (e.g., sampling algorithms, data collection interventions). However, researchers do not always motivate why they measure a data diversity dimension in a certain way.

\paragraph*{A broader view: Components of diversity dimensions} 
Data diversity dimensions can be decomposed into different components.
Ecology, with its long tradition of diversity measurement, often recognizes two core components of biodiversity: richness (i.e., the number of species) and evenness (i.e., how balanced a distribution of species is) \citep{magurran2004measuring}. Other components are sometimes also considered, like disparity (taking the similarity between species into account) \citep{math6070119}. While ideas from ecology may not map perfectly onto NLP, especially for dimensions that are not easily captured by categorical data, it offers a useful lens for thinking about data diversity \citep{friedman2023vendi,esteve2025surveydiversityquantificationnatural,lion-bouton-etal-2022-evaluating}. 
For instance,  in language learning, ideas from ecology have been applied to measuring lexical diversity \citep{scott2013}. Similarly, for topical diversity, we could consider the number of distinct topics (richness), how the data is distributed over topics (evenness), and the semantic similarity between topics (disparity).

 NLP researchers, however, are rarely explicit about which components of a diversity dimension are considered. This ambiguity can lead to inconsistent or misleading interpretations. For example, does \textit{increasing dataset size
lead to greater diversity}?
If the diversity measure captures richness then adding more data generally helps---or at least doesn’t hurt. However, if a balanced distribution across categories (i.e., evenness) is targeted, simply scaling up the dataset may not improve diversity and could even reinforce imbalances.

\paragraph*{Better data diversity measures: Be critical of representations}
Many measures rely on mapping data to some representation, e.g., discrete categories or continuous embeddings. 
However, the choice of representation shapes what is visible and measurable. 
For example, consider measuring semantic diversity. The quality of embeddings may vary across topics or dialects,
and they may capture information (e.g., style, \citealt{wegmann-etal-2022-author}) beyond semantics. 
Furthermore, measuring diversity along social dimensions (e.g., annotator background) often involves categorizing people. However, attributes like gender, race and ethnicity  are socially constructed and often oversimplified in datasets \citep{10.1145/3630106.3659050,doi:10.1146/annurev-soc-030420-015435}. Social science research can offer useful insights and tools for more nuanced representations and measurement \citep{10.1145/3630106.3659050,doi:10.1146/annurev-soc-030420-015435}.

\paragraph*{A broader view: Beware of cross-lingual differences}
Diversity measures often rely on specific tools or models (e.g., tokenizers, embedding models), and the availability and quality of such tools varies between languages \citep{blasi-etal-2022-systematic}. Some measures cannot be applied to all languages. For example, measuring  syntactic diversity using a dependency parser \citep{guo-etal-2024-curious} is not possible if such a tool is lacking.
Even when tools are available, the quality of diversity measurements can vary across languages.
For example, \citet{lion-bouton-etal-2022-evaluating} showed that errors in automatic lemmatization can artificially inflate diversity scores.
Structural properties of languages can also influence the distribution of diversity scores.
For instance, type-token ratio (TTR), often used to measure lexical diversity, was found to correlate with morphological complexity \citep{kettunen2014can}.
Therefore, special care should be taken when applying and comparing diversity measures across languages.

\subsection{Desiderata and Evaluation}
\label{sec:evaluation}

What makes a good diversity measure? This question is rarely addressed, and diversity measures are rarely evaluated (with a few exceptions; \citealt{tevet-berant-2021-evaluating,han-etal-2022-measuring,yang2025measuringdatadiversityinstruction,zhang-etal-2025-evaluating-evaluation}). 
Yet without doing so, it is unclear whether a measure  captures what is intended or improves on others. 
 
\paragraph*{Better and broader evaluation: Leveraging measurement theory} 
In the social sciences, measurement theory is widely used to evaluate the quality of measurements for abstract concepts, like personality or self-esteem \citep{trochim2016research}. 
Its benefits have already been highlighted for
measuring biases \citep{van2024undesirable} and evaluating AI systems \citep{wallach2025positionevaluatinggenerativeai,fang-etal-2025-patch}.
It also holds potential for data diversity measurement \citep{icml2024_position_paper}.
Two key indicators of measurement quality are (construct) validity (i.e., are we measuring what we intend to measure) and reliability 
(i.e., the consistency or repeatability of the measurements). Although measurement theory provides a useful toolbox, translating its methods and concepts to NLP also poses challenges \citep{van2024undesirable}.

To illustrate how measurement theory can inform the evaluation of a diversity measure's validity, consider the following three complementary sources of construct validity evidence \citep{trochim2016research}. Content validity involves assessing whether the full scope of a concept is adequately covered. Suppose we aim to measure dialect diversity in a dataset: if the chosen measure covers only a small subset of major dialects, this could point to a content validity problem. Convergent validity is another example, which examines whether multiple measures for the same diversity dimension produce similar diversity scores \citep{shaib2025standardizingmeasurementtextdiversity}. 
Finally, criterion validity tests whether the measure behaves as expected according to an external benchmark---for example, by constructing datasets with known differences in diversity and verifying that the measure correctly ranks them. Together, these three validity indicators provide complementary evidence of a measure's validity.

\paragraph*{Better data diversity measures: Defining desiderata}
 The NLP community should also engage more deeply with the question of what makes a good diversity measure. Desiderata can be theoretical, for example that a measure should be continuous or clarity on the conditions under which it is maximal \citep{math6070119}. Desiderata can also be practical like the interpretability or actionability of measures \citep{delobelle-etal-2024-metrics}. For example, measures based on pairwise similarities between embeddings are increasingly popular. These measures are often transparent in their formulation, but their dependence on embedding models complicates interpretation; furthermore, while they indicate how similar instances are, they do not indicate what is missing from a dataset, making it more difficult to intervene.
The  efficiency of a diversity metric may also be an important practical desideratum. \citet{shaib2025standardizingmeasurementtextdiversity} point out that some metrics  are prohibitively slow and therefore not suitable for large-scale application.

\section{Conclusion}
\label{sec:conclusion}
We have examined key challenges in measuring data diversity in NLP and conclude with six immediately actionable recommendations.
These recommendations relate to \textit{what} diversity aspects should be measured (\S\ref{sec:dimensions}), \textit{how} they should be measured (\S\ref{sec:stat_properties}), and the \textit{assessment} of such measures (\S\ref{sec:evaluation}):
(i) Clearly define what is meant by data diversity in the context of a study;
(ii) Make deliberate, transparent choices about how diversity is measured. They should be guided by the specific research goals and application;
(iii) Develop measures that better capture the social dimensions of diversity;
(iv) Critically examine what makes a good diversity measure. NLP needs clearer guidance on the desirable theoretical and practical properties of data diversity measures;
(v) Evaluate the quality of data diversity measurement; and finally
(vi) Look beyond NLP. Many fields have developed approaches to measure diversity; we should learn from them.

\section*{Limitations}
Diversity is a rich and complex concept. Although we take the view that researchers should measure a dataset's diversity, we acknowledge that any quantification necessarily results in a loss of nuance and information.
Moreover, not all dimensions of diversity are necessarily measurable.
Engaging with qualitative perspectives on diversity is thus important; it could bring a critical and context-sensitive lens to    diversity
in NLP datasets. We also believe that combining manual and automatic analyses of datasets is an important path forward \citep{prabhu2020largeimagedatasetspyrrhic}.
Relatedly, we acknowledge that quantifying data diversity may lead to diversity metrics becoming objectives on their own.
We caution against the blind optimization of diversity metrics, as the quality of a dataset depends on various factors. Purely optimizing for a diversity metric might overlook other important factors of dataset quality, such as the relevance of texts.

\section*{Acknowledgments}
We would like to thank the members of the NLP and Society Lab at Utrecht University and the anonymous reviewers for their feedback.
This work is supported by the ERC Starting Grant DataDivers 101162980 and the Carlsberg Foundation under the \textit{Semper Ardens: Accelerate} programme (CF21-0454).
We have used ChatGPT to improve the grammar and phrasing of some sentences in this paper.

\bibliography{anthology,custom}

\end{document}